\documentclass{article} 
\usepackage{times}

\usepackage{amsmath,amsfonts,bm}

\def\figref#1{figure~\ref{#1}}

\def\secref#1{section~\ref{#1}}

\def\eqref#1{equation~\ref{#1}}

\def\1{\bm{1}}

\DeclareMathAlphabet{\mathsfit}{\encodingdefault}{\sfdefault}{m}{sl}
\SetMathAlphabet{\mathsfit}{bold}{\encodingdefault}{\sfdefault}{bx}{n}

\usepackage{float}
\usepackage[utf8]{inputenc} 
\usepackage[T1]{fontenc}    
\usepackage{microtype}      
\usepackage{color,graphicx}
\usepackage{url}            
\usepackage{booktabs}       
\usepackage{hyperref}       
\usepackage{amsfonts}       
\usepackage{nicefrac}       
\usepackage{amsmath}
\usepackage{subcaption}

\usepackage[accepted]{icml2019}

\newcommand{\TODO}[1]{\ifthenelse{\boolean{include-notes}}
 {{\color{red} TODO: #1}}{}}
\newcommand{\Coment}[1]{\ifthenelse{\boolean{include-notes}}
 {{\color{blue} #1}}{}}
\newcommand{\noah}[1]{\ifthenelse{\boolean{include-notes}}
 {{\color{green}Noah: #1}}{}}

\newboolean{include-notes}
\setboolean{include-notes}{false}

\newcommand{\adnote}[1]{\ifthenelse{\boolean{include-notes}}
 {\textcolor{red}{\textbf{A:#1}}}{}}

\newcommand{\algoref}[1]{Algorithm~\ref{#1}}
\newcommand{\eqnref}[1]{Equation~\ref{#1}}
\newcommand{\tblref}[1]{Table~\ref{#1}}
\newcommand{\angles}[1]{\langle #1 \rangle}

\usepackage[font=footnotesize,labelfont=bf]{caption}
\newcommand{\prg}[1]{\noindent\textbf{#1}}

\icmltitlerunning{On the Feasibility of Learning Biases for Reward Inference}

\begin{document}

\twocolumn[
\icmltitle{On the Feasibility of Learning, Rather than Assuming, \\ Human Biases for Reward Inference}




\begin{icmlauthorlist}
\icmlauthor{Rohin Shah}{berkeley}
\icmlauthor{Noah Gundotra}{berkeley}
\icmlauthor{Pieter Abbeel}{berkeley}
\icmlauthor{Anca D. Dragan}{berkeley}
\end{icmlauthorlist}

\icmlaffiliation{berkeley}{Department of Electrical Engineering and Computer Science, UC Berkeley}

\icmlcorrespondingauthor{Rohin Shah}{rohinmshah@berkeley.edu}

\icmlkeywords{Preference learning, inverse reinforcement learning}

\vskip 0.3in
]

\printAffiliationsAndNotice{}

\begin{abstract}


Our goal is for agents to optimize the \emph{right} reward function, despite how difficult it is for us to specify what that is. Inverse Reinforcement Learning (IRL) enables us to infer reward functions from demonstrations, but it usually assumes that the expert is noisily optimal. Real people, on the other hand, often have \emph{systematic} biases: risk-aversion, myopia, etc. One option is to try to characterize these biases and account for them explicitly during learning. But in the era of deep learning, a natural suggestion researchers make is to avoid mathematical models of human behavior that are fraught with specific assumptions, and instead use a purely data-driven approach. We decided to put this to the test -- rather than relying on assumptions about which specific bias the demonstrator has when planning, we instead learn the demonstrator's \emph{planning algorithm} that they use to generate demonstrations, as a differentiable planner. Our exploration yielded mixed findings: on the one hand, learning the planner can lead to better reward inference than relying on the wrong assumption; on the other hand, this benefit is dwarfed by the loss we incur by going from an exact to a differentiable planner. This suggest that at least for the foreseeable future, agents need a middle ground between the flexibility of data-driven methods and the useful bias of known human biases. Code is available at \url{https://tinyurl.com/learningbiases}.

\end{abstract}




\section{Introduction}

\begin{figure}
    \centering
    \includegraphics[width=\columnwidth]{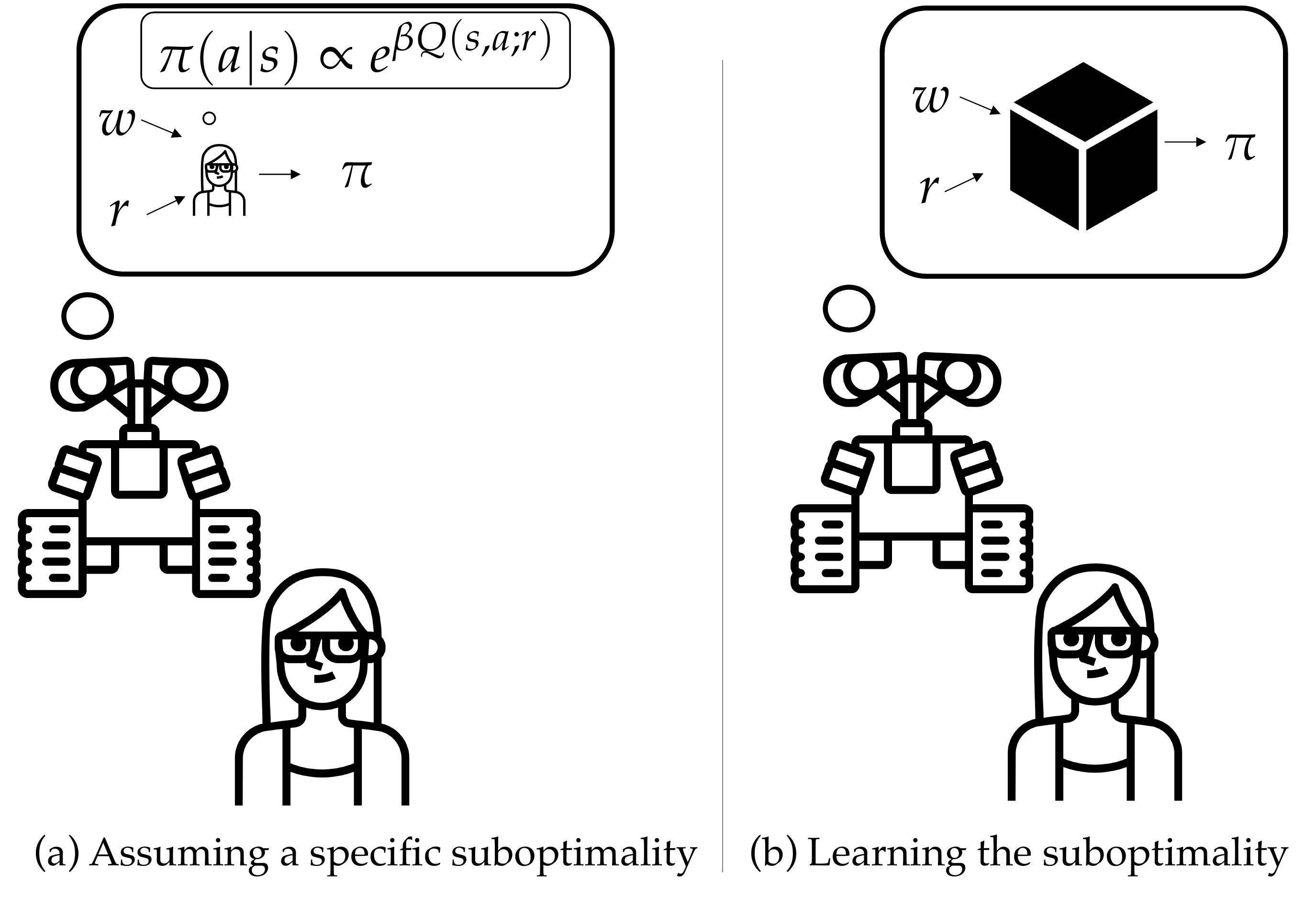}
    \caption{While we could correct for systematic biases by having our AI system reason about explicit models of human reasoning, using the wrong assumption can lead to agents that do not correctly understand what people want. A natural alternative is to learn human biases from data. Our goal in this work is to investigate this alternative and gain insight into what additional assumptions might make it feasible and what algorithmic improvements it demands.}
    \label{fig:front}
\end{figure}


Our ultimate goal is to enable the design of agents that optimize for the \emph{right} reward function. Unfortunately, designing reward functions is challenging \citep{DeepRLHumanBlog} and can have unintended side-effects \citep{IRD,SpecificationGaming}. Inverse Reinforcement Learning (IRL) \citep{IRLIntro,IRL2000,IRL2004} aims to bypass the need for reward design by learning the reward from observed demonstrations of good behavior.

Existing IRL algorithms typically make the assumption that the demonstrator is either optimal, or Boltzmann rational, i.e. taking better actions with higher probability \citep{MaxentIRL,GCL}. However, there is a rich literature showing that humans are \emph{not} optimal, and are biased in \emph{systematic} ways. Consider a grad student who starts writing a paper a month in advance, expecting it to take two weeks, but then misses the deadline. Should we infer that they prefer to lose sleep to pursue a deadline that they then miss? Of course not. This is a classic case of the \emph{planning fallacy} \citep{PlanningFallacy}: the grad student was wrong in their prediction of how long it would take to complete the paper. An assumption of noisy rationality cannot allow us to correct for this bias: how do we tell whether the grad student underestimated the time required and actually wanted to finish the paper earlier, rather than overestimating the time required and actually wanting to finish the paper even later?



Of course, if we know that humans tend to underestimate how long any given task will take, then we can correct for this bias by having our AI system reason about how it affects human reasoning, as illustrated in \figref{fig:front}a. IRL algorithms have been developed that can account for particular systematic biases, such as myopia and hyperbolic time discounting \citep{IgnorantInconsistentAgents,BoundedAgents}, sparse noise \citep{SparseNoiseIRL}, risk sensitivity \citep{RiskIRL}, or a bad dynamics model \citep{InferringDynamics}. Even suboptimal trajectories or failures \citep{IRLFromFailure} can be thought of as a biased demonstrator, where the bias is the specific model of failure. However, choosing a particular model of suboptimality is a big assumption, and can lead to arbitrarily bad performance if the assumption is incorrect \citep{ModelMisspec,ModelMisspecIRL}. For example, if we try to explain the grad student's behavior as hyperbolic time discounting (that is, valuing short-term rewards disproportionately more than long-term ones), we might infer that the grad student enjoys long nights of writing over the short term, rather than viewing it as an instrumental goal necessary for submitting the paper. We wouldn't want our AI system deleting our in-progress paper so that we can have the ``joy'' of rewriting it from scratch!



Given how complex real humans are, it seems hopeless to know exactly which bias a person is displaying, and inevitable that any such assumption will lead to a misspecified bias model. In the age of data-driven methods, it seems almost natural to think that we could learn the bias model as well, rather than relying on a hardcoded assumption. One can view the demonstrator's behavior as a composition of a reward function and a \emph{planning algorithm} that computes what actions to take given a reward function -- instead of assuming a planning algorithm (e.g. noisy rational, myopic, etc.), why not learn it? Our work is about exploring the feasibility of this alternative (depicted in \figref{fig:front}b). 


Right off the bat, this enticing and seemingly natural idea hits a wall: unfortunately, when the planning algorithm can be any function mapping reward functions to policies, it is impossible to learn the true reward function even with infinite data, because there are always alternative explanations for the observed policy \citep{ImpossibleIRL,EasyGoalInference}. Any particular behavior could be explained either by positing a term in the reward function, or a bias in the planning algorithm. Rather than using data to avoid all assumptions, our work actually investigates whether it is at least feasible to learn the planning algorithm when either a) we get to first observe demonstrations in tasks where we know the reward, and can thus focus only on learning the planner for those demonstrations; or b) we assume that the demonstrator is good at the tasks, and regularize the planning algorithm towards optimality. 
We find that even in the ``easy'' setting where we are given access to some rewards, this problem is very difficult: there is some benefit from learning systematic biases in the planner, but this is dwarfed by the disadvantage of using a neural network as a planner, relative to using an exact, perhaps slightly misspecified model of rationality. In the case where we instead regularize towards optimality, the resulting algorithm is more robust to different human models, but again the benefit is dwarfed by the inaccuracies introduced by differentiable planning.

\section{Examples of Biases}
To put this work in context, we start with some examples of the kind of biases a general algorithm should be able to capture and account for. While we use these for illustrative purposes, the whole point of our exploration is that humans might have systematic suboptimalities completely different from these examples. We don't know all the possible biases a priori -- if we did, using that information would certainly be the superior choice.

\begin{figure*}
    \centering
    \includegraphics[width=\linewidth]{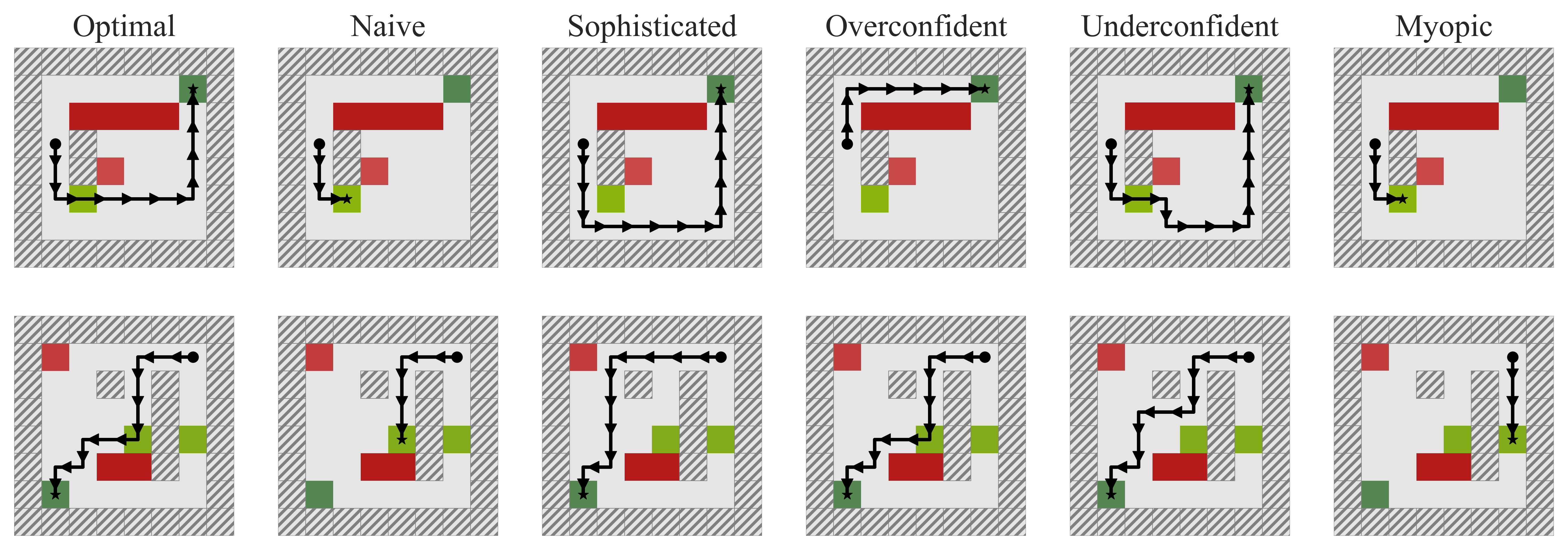}
    \caption{The plans of our synthetic agents on two navigation environments. Actual trajectories could differ due to randomness in the transitions. Green squares indicate positive reward while red squares indicate negative reward, with darker colors indicating higher magnitude of reward.}
    \label{fig:biases}
\end{figure*}

\prg{Running Example.} We illustrate the effects of these biases on a simple 2D navigation task in \figref{fig:biases}. There are multiple salient locations, each of which has a desirability score (which can be negative, in which case the agent wants to avoid those locations). The agent can move in any of the four cardinal directions, or stay in its current position. Every movement action has a chance of failing and causing the agent to move in a direction orthogonal to the one it chose. Despite their simplicity, there are several ways in which human-like suboptimal behavior can manifest in these environments.

\prg{Time inconsistency.}
Would you prefer to get \$100 in 30 days, or \$110 in 31 days? Faced with this question, people typically choose the latter. However, thirty days later, when faced with the choice of getting \$100 now, or \$110 tomorrow, they sometimes choose to take the \$100. This reversal of preferences over time would never happen with an optimal agent that maximizes expected sum of discounted rewards. Researchers model this phenomenon using \emph{hyperbolic time discounting}, in which future rewards are discounted more aggressively than exponentially. This leads to a followup question -- how do humans make long-term plans, given that their future self will have different preferences? Prior work has considered a spectrum from \emph{naive} agents that assume their future self will have the same preferences as they do, to \emph{sophisticated}  agents that perfectly understand how their preferences will change over time and make plans that take such change into account \citep{HyperbolicDiscounting}.

In \figref{fig:biases}, when going to a high reward, both the \emph{naive} and \emph{sophisticated} hyperbolic time discounters can be "tempted" by a proximate smaller reward. The naive agent fails to anticipate the temptation, and so once it gets near the smaller positive reward, it caves in to the temptation and stays there. The sophisticated agent explicitly plans to avoid the temptation -- it does not collect the smaller reward and instead takes a longer, more dangerous path around the smaller reward to get to the large reward.

\prg{Incorrect estimates of probabilities.}
Humans are notoriously bad at judging probabilities. The \emph{availability heuristic} \citep{AvailabilityHeuristic} refers to the human tendency to rate events as more likely if they are easier to recall. The \emph{recency effect} is a similar effect where recent events are judged to be more probable. These biases are depend heavily enough on context that they don't transfer to our task in any obvious way. So, we use two simplified models -- an \emph{overconfident} agent, which expects that the most likely next state is more likely than it actually is, leading it to take what we would call risky behavior, and an \emph{underconfident} agent, which analogously behaves in an overly cautious manner. In \figref{fig:biases}, the overconfident agent takes the shortest path to the reward, underestimating the risk of slipping into the large region of negative reward, while the underconfident agent plans a circuitous route around negative reward that it is unlikely to have actually encountered.

\prg{Bounded computation.}
Researchers have studied models of \emph{bounded rationality}, where humans are assumed to be rational subject to the constraint that they have a bounded amount of computation. This can be thought of as an explanation that many other heuristics and biases are actually computational shortcuts that allow us to reach reasonably good decisions without too much cost \citep{BoundedRationality}. In our task, we model computation bounds as a small time horizon for planning, leading to \emph{myopic} behavior. In \figref{fig:biases}, the myopic agent can only see close rewards, and goes directly to them, never even realizing the possibility of going to the highest reward.

\section{Problem: Learning Rewards of Demonstrators with Unknown Biases}

\noindent\textbf{Notation.} A (finite-horizon) Markov Decision Process (MDP) \citep{MDPDefinition} is a tuple $\angles{S, A, T, r, H}$. $S$ is a set of states. $A$ is a set of actions. $T$ is a probability distribution over the next state, given the previous state and action. We write this as $T(s_{t+1} | s_t, a)$. $r$ is a reward function that maps states and actions to rewards $r : S \times A \rightarrow \mathbb{R}$. $H \in \mathbb{Z}_{+}$ is the finite planning horizon for the agent. Since we are interested in the setting where the reward function $r$ is unknown, we will factor MDPs into \emph{world models} $w=\angles{S, A, T, H}$ and reward functions $r$.

Instead of having access to a reward function, we observe the behavior of a demonstrator, who performs the task well but could be suboptimal in systematic ways. We assume that the demonstrator produces (possibly stochastic) policies using a \emph{planning algorithm} $D : (\mathbb{W} \times R) \rightarrow (S \rightarrow A \rightarrow [0, 1])$, or planner for short. Here $\mathbb{W}$ is a space of world models with the same set of states $S$ and actions $A$, and $R \subseteq S \times A \rightarrow \mathbb{R}$ is a space of reward functions that the demonstrator can plan for. Later in this section we illustrate additional assumptions about $D$. We observe the demonstrator's \emph{policy} $\pi : S \rightarrow A$ for a particular world model $w$, with $\pi_D = D(w, r^*)$ for some unknown reward $r^*$.

\noindent\textbf{Estimating Biases and Rewards.} Given a \textbf{\emph{world model}} $w$ and the \textbf{\emph{demonstrator's policy}} $\pi_D$ which may exhibit an unknown bias, determine the \textbf{\emph{reward $r^*$}} that the demonstrator is optimizing.

We might hope that enough data can solve this problem without any additional assumptions. However, this problem is unsolvable -- \citet{ImpossibleIRL} prove an impossibility result showing that for any potential reward function $r'$, there is some planner $D'$ such that $D'(w, r') = D(w, r^*)$. The proof is simple -- simply set $D'(w, r') = \pi_D$ for any $r'$, that is $D'$ always returns $\pi_D$ regardless of the reward function. What we really explore in this work is thus data-driven approaches that make minimal additional assumptions, rather than none at all.

Inverse reinforcement learning assumes that the demonstrator is (approximately) optimal to get around this issue. It is common to assume Boltzmann rationality, where the probability of an action is proportional to the exponent of its expected value \citep{BoltzmannQvals}, i.e. $P(a | s) \propto e^{Q(s, a)}$, where $Q$ is the optimal Q function that satisfies the Bellman equation:
\begin{equation}
    Q(s, a) = r(s, a) + \gamma \sum\limits_{s'} \left[ T(s' | s, a) \max\limits_{a'} Q(s', a') \right]
\end{equation}

However, we know that humans are systematically suboptimal, and so we would like to relax this assumption and try other, more realistic assumptions. The pathological solutions in the impossibility result occur partly because the demonstrator can have arbitrary behavior on different environments. While we certainly want the demonstrator to adapt to different environments, the \emph{algorithm} that the demonstrator uses to determine their policy should stay fixed across similar environments. This imposes structure on the demonstrator's planner that can eliminate some possibilities.

\noindent\textbf{Assumption 1:} The demonstrator plans in the same way for sufficiently similar environments.

Intuitively, the demonstrator's planning algorithm $D : (\mathbb{W} \times R) \rightarrow (S \rightarrow A \rightarrow [0, 1])$ is ``the same'' for similar environments. Of course, if $D$ can be any function with this type signature, it can still map any arbitrary $(w, r)$ pair to any arbitrary policy $D(w, r)$, but we will further ensure that $D$ is simple (through regularization). Given a list of world models $W = [w_1 \dots w_n]$ and reward functions $R = [r_1 \dots r_n]$, we define $D(W, R)$ to be the list of the demonstrator's policies $[D(w_1, r_1) \dots D(w_n, r_n)]$.

Note that this is a strong assumption: while it is reasonable to believe that people plan in the same way for variations of the same task, they likely have different biases for different tasks, because they may have domain-specific heuristics. The setting of multiple tasks has been studied before \citep{MultitaskMaxEntIRL,BayesianMultitaskIRL,NonparametricMultitaskIRL,MetaIRL}, though not for the purpose of inferring systematic biases.

This assumption leads to a slightly easier problem, of recovering rewards from multiple tasks:

\noindent\textbf{Estimating Biases and Rewards for Multiple Tasks.} Given a list of world models $W$ and the demonstrator's policies $\Pi_D = D(W, R)$ which may exhibit an unknown bias, determine the list of reward functions $R$ (one for each $w \in W$) that the demonstrator was optimizing.

Since the person uses the same planner across all tasks, an agent can have an easier time recovering rewards for each task by leveraging the common structure across the tasks. This is especially appealing for agents that would get to observe people for some period of time before trying to assist them. However, Assumption 1 still falls prey to the impossibility result. Consider the case where the demonstrator is optimal. Given the assumptions so far, we could infer that the demonstrator is \emph{minimizing} expected reward for the reward function $-r^*$, since that perfectly predicts $\pi_D$. This is very bad, as we could infer a reward that incentivizes the \emph{worst} possible behavior!

When humans take action, we typically assume that they are doing something that is reasonable for achieving their goals, even if it is not optimal:

\prg{Assumption 2a:} The demonstrator is ``close'' to optimal.

This is a weaker version of the standard IRL assumption of Boltzmann rationality. In \secref{sec:em_algorithm}, we derive a natural algorithm that takes advantage of this assumption, by regularizing the planner towards optimality. This gives us an algorithm for the problem of estimating biases and rewards for multiple tasks that does not obviously fail due to an impossibility result.

We also explore an alternative approach, based on the fact that we have strong priors about what humans are trying to optimize for. Intuitively, these priors allow us to infer how good they are at achieving their goals, and in what ways they are systematically biased, which can be used to better infer goals in new settings. We formalize this by assuming that we observe some tasks where we know the demonstrator's reward function and policy.

This again gives us a natural algorithm, detailed in \secref{sec:known-r-alg} that does not obviously fail due to the impossibility result, since we can easily infer that the demonstrator is ``close'' to optimal from the tasks for which we do observe the demonstrator's reward function.

\noindent\textbf{Assumption 2b:} We know what reward function the demonstrator is optimizing for \emph{some} tasks.

\noindent\textbf{Estimating Biases and Rewards with Access to Tasks with Known Rewards.} Given a list of world models $W$, a list of the demonstrator's policies $\Pi_D = D(W, R)$, a list of world models $W_{\text{known}}$ with known rewards $R_{\text{known}}$ and a list of the demonstrator's policies $\Pi_{\text{known}} = D(W_{\text{known}}, R_{\text{known}})$, determine the reward functions $R$ that $D$ was optimizing.

While our problem formulations above assume that we have access to full policies $\Pi_D$, none of the algorithms rely on this assumption -- it is easy to modify them to work with trajectories instead.

\section{Algorithms to Estimate Biases and Rewards}

The idea that we investigate in this work is whether it is beneficial to learn a model of how the demonstrator plans. Once we have learned the planning algorithm $D$, we are faced with an inverse problem: we want to find the $r$ such that $\pi_D = D(w, r)$. This resembles the problem of feature visualization for image classifiers \citep{FeatureVisualization}, and suggests a natural approach: as long as the planner $D$ is differentiable, we can invert its ``understanding'' using backpropagation to infer the reward from the policy.

\subsection{Architecture}

We model the demonstrator planning algorithm $D$ using a \emph{differentiable planner} $f_{\theta}$, which is a neural net that can express planning algorithms whose parameters $\theta$ can be updated using gradient descent. $f$ has the same type as the demonstrator's planner $D$, namely $(\mathbb{W} \times R) \rightarrow (S \rightarrow A \rightarrow [0, 1])$. Thus, the inputs to the differentiable planner $f$ are a world model $w \in \mathbb{W}$ and a reward function $r \in \mathbb{R}$; the output is a stochastic policy $\pi \in (S \rightarrow A \rightarrow [0, 1])$. We determine how well $\angles{f, R}$ matches the demonstrator's policy $\pi_D$ with the cross entropy loss $\mathcal{L}(f_{\theta}(W, R), \Pi_D)=\sum\limits_i \mathcal{L}(f_{\theta}(w_i, r_i), \pi_{D,i})$.

\begin{figure*}
\centering
\begin{subfigure}{.48\textwidth}
  \centering
  
  \includegraphics[width=\linewidth]{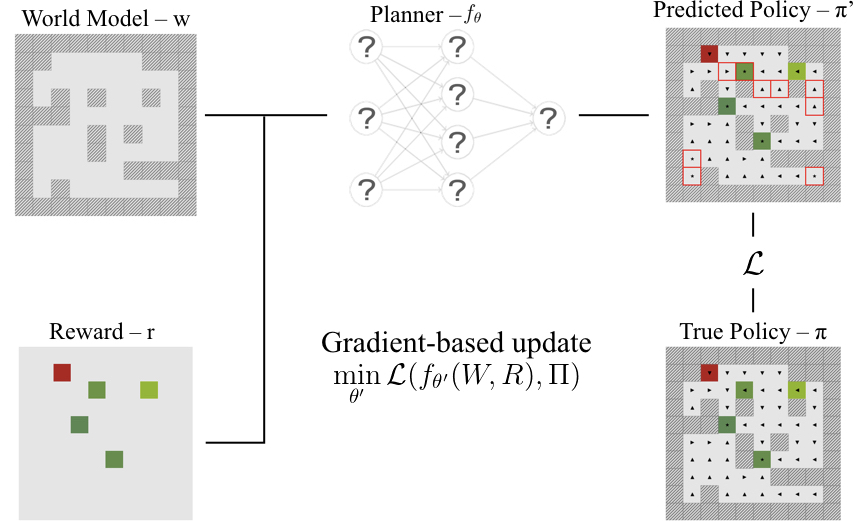}
  \caption{Training the planner $f_{\theta}$. We hold the world model $w$, reward $r$, and policy $\pi$ fixed, and update $f_{\theta}$ with gradient descent.}
  \label{fig:train-planner}
\end{subfigure} \space \space
\begin{subfigure}{.48\textwidth}
  \centering
  \includegraphics[width=\linewidth,keepaspectratio]{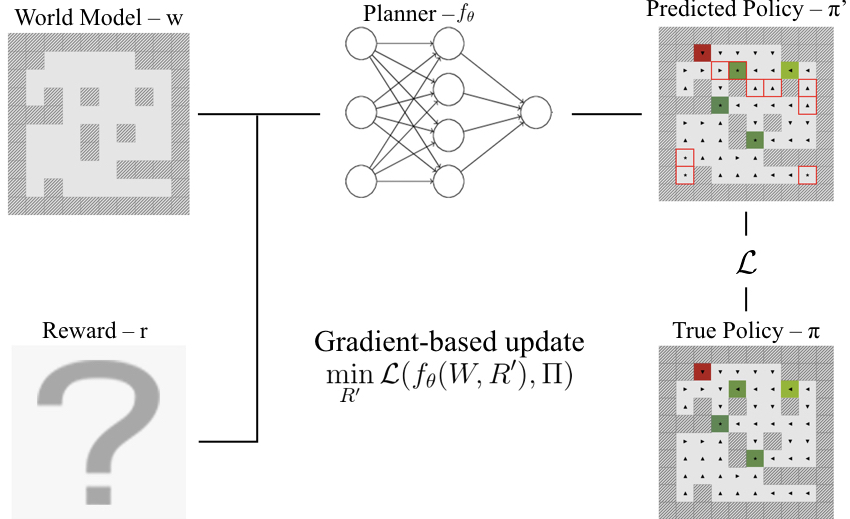}
  \caption{Training the reward $R$. We hold the world model $w$, planner $f_{\theta}$, and policy $\pi$ fixed, and update the reward $r$ with gradient descent.}
  \label{fig:train-reward}
\end{subfigure}
\setlength{\abovecaptionskip}{15pt plus 3pt minus 3pt}
\caption{The architecture and operations on it used in the algorithms.}
\label{fig:architecture}
\end{figure*}

While the algorithms can work with any differentiable planner, in this work we use a \emph{value iteration network} (VIN) \citep{VINs}. A VIN is a fully differentiable neural network that embeds an approximate value iteration algorithm inside a feed-forward classification network. For environments where transitions only depend on "nearby" states (as in navigation tasks), the Bellman update can be performed using an appropriate convolution, and the computation of values from Q-values can be done with a max-pooling layer. By leaving the filters for the convolutions unspecified, the VIN can automatically learn the transition probabilities. Of course, the VIN is merely one architecture for a differentiable planner; we could equally well use other planners \citep{UPNs,ModelBasedPlanner,MCTSNets}. The algorithms we study will become stronger as research in this area advances.




\prg{The components of the algorithms.} This architecture enables two main operations that are important for inferring rewards and biases, which we illustrate in \figref{fig:architecture}. First, given world models $W$, reward functions $R$ (either known or hypothesized), and the demonstrator's policies $\Pi_D$, we can train a corresponding planner using gradient descent (\figref{fig:train-planner}):

\begin{equation}
\tag{\textsc{Train-Planner}}
\theta = \min_{\theta'} \mathcal{L}(f_{\theta'}(W, R), \Pi_D)
\end{equation}

Second, given world models $W$, demonstrator's policies $\Pi_D$, and some planner parameters $\theta$, we can infer the corresponding reward functions using gradient descent (\figref{fig:train-reward}):

\begin{equation}
\tag{\textsc{Train-Reward}}
R = \min_{R'} \mathcal{L}(f_{\theta}(W, R'), \Pi_D)    
\end{equation}

It is also possible to perform both of these at the same time by training the planner parameters and rewards jointly given world models $W$ and the demonstrator's policies $\Pi_D$:

\begin{equation} \label{eq:joint}
\tag{\textsc{Train-Jointly}}
R, \theta = \min_{R', \theta'} \mathcal{L}(f_{\theta'}(W, R'), \Pi_D)
\end{equation}

\subsection{Learning the planner from known rewards first (Assumption 2b)} \label{sec:known-r-alg}

Consider the simpler setting when we have access to a set of tasks with \emph{known} rewards. The known rewards can be used to infer the planning algorithm used by the demonstrator, which can then be used to infer rewards in the remaining cases. So, the planner is first trained on the world models for which we have rewards. Then, learned planner weights allow us to infer the reward on the world models for which we don't know the reward. This algorithm is illustrated in Algorithm~\ref{alg:easy}. 

\begin{algorithm}
\begin{algorithmic}[1]
    \STATE $\theta \Leftarrow$ \textsc{Train-Planner}$(W_{\text{known}}, R_{\text{known}}, \Pi_{\text{known}})$
    \RETURN \textsc{Train-Reward}$(W, \theta, \Pi_D)$
\end{algorithmic}
\caption{\textsc{irl-with-rewards}: Estimating biases and rewards with access to tasks with known rewards. Requires that Assumptions 1 and 2b hold.}
\label{alg:easy}
\end{algorithm}

\subsection{Learning the planner and rewards simultaneously (Assumption 2a)} \label{sec:em_algorithm}

Perhaps the most natural algorithm to solve our problem would be to jointly train the planner and rewards on the given set of world models and policies. However, this falls prey to the impossibility result: there is no way to distinguish between reward maximization with an optimal reward $r^*$ and reward minimization with the reward $-r^*$. So, it would be useful to regularize the planner so that it is ``close'' to optimal, in accordance with Assumption 2a.

A natural way to do this regularization is by \emph{initializing} the planner to be optimal, and then finetuning the result by training jointly as in \eqnref{eq:joint}. Since the reward inference requires a differentiable planner, we need a method that sets a differentiable planner to be optimal (that is, the planner that maximizes expected reward). This can be done by simulating data from an optimal agent with randomly generated world models and rewards, and use this to train the planner to mimic an optimal agent. The resulting algorithm is illustrated in \algoref{alg:hard}. We show in \secref{sec:initialization} that the initialization is crucial for good performance as we would expect.

\begin{algorithm}
\begin{algorithmic}[1]
    \STATE $W_{\text{sim}}, R_{\text{sim}} \Leftarrow$ Generate world models and rewards
    \STATE $\Pi_{\text{sim}} \Leftarrow$ Run optimal agent on $\angles{W_{\text{sim}}, R_{\text{sim}}}$
    \STATE $\theta_{\text{init}} \Leftarrow$ \textsc{Train-Planner}$(W_{\text{sim}}, R_{\text{sim}}, \Pi_{\text{sim}})$
    \STATE $R_{\text{init}} \Leftarrow$ \textsc{Train-Reward}$(W, \theta_{\text{init}}, \Pi_D)$
    \STATE $\theta, R \Leftarrow$ \textsc{Train-Jointly}$(W, \Pi_D)$  \COMMENT{using $\theta_{\text{init}}$ and $R_{\text{init}}$ as initializations}
    \RETURN $R$
\end{algorithmic}
\caption{\textsc{irl-without-rewards}: Estimating biases and rewards for multiple tasks with no known rewards. Requires that Assumptions 1 and 2a hold.}
\label{alg:hard}
\end{algorithm}

\section{Evaluation} \label{sec:evaluation}

We evaluate the algorithms by simulating demonstrators with different biases, and testing whether the same method can correctly infer reward for all these demonstrators.

\subsection{Experiment details}

In all experiments below, results are averaged over 10 runs with different seeds, on randomly generated 14x14 gridworlds that have 7 squares with non-zero rewards. We ensure that all such squares can be reached from the start state, and that at least half of the positions in grid are not walls.

We use a Value Iteration Network with 10 iterations as the differentiable planner, and set the space of rewards to be $S \rightarrow \mathbb{R}$; that is, any state can be mapped to any reward, but the reward is assumed not to depend on the action. We added an extra convolutional layer to the initial part of the VIN (which learns a proxy reward) as initial experiments showed that this could better learn an optimal planner for our gridworlds; other than that the architecture remains as described in \citet{VINs}. We apply L2 regularization to the VIN with scale 0.0001, and do not regularize the reward.

For all experiments, we kept the number of demonstrations fixed to 8000. For Algorithm 1, this was split into 7000 policies with rewards that were used to train the planner, and 1000 on which rewards had to be inferred. Note that this does \emph{not} include any simulated data -- for example, Algorithm 2 would get 8000 biased policies, and would \emph{also} simulate a further 7000 policies from an optimal agent in order to initialize the planner and reward.

\subsection{Evaluating reward inference}

\begin{figure*}
    \centering
    \includegraphics[width=\linewidth]{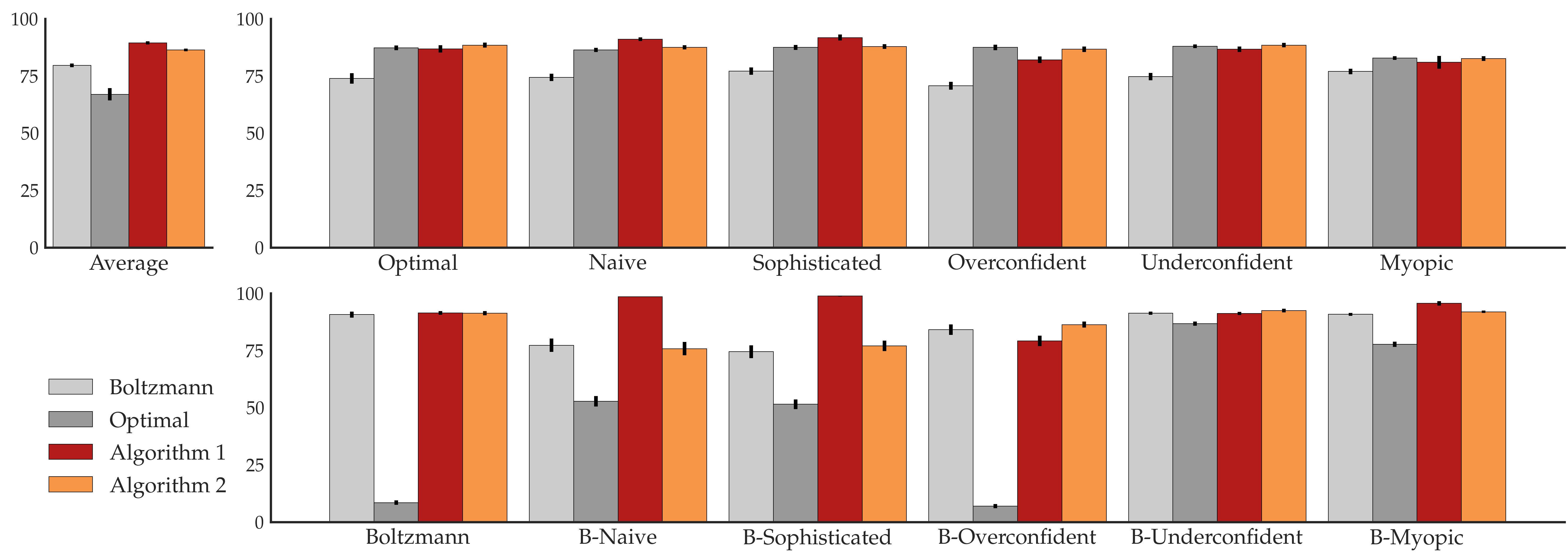}
    \caption{Reward obtained when planning with the inferred reward, as a percentage of the maximum possible reward, for different bias models and algorithms. We implement five types of biases by modifying the value iteration algorithm to produce a different set of Q-values. The top row shows results for agents that choose between the best actions from these Q-values, while in the bottom row the agent chooses actions with probability proportional to the exponent of the Q-value (the Boltzmann assumption). The average over both rows is displayed in the top left.}
    \label{fig:bias-results}
\end{figure*}

\prg{Hypothesis.}
The hypothesis we put to the test in this work is that accounting for unknown systematic bias should outperform the assumption of a particular inaccurate bias, e.g. noisy rationality or the lack thereof. 

\prg{Manipulated variables.}
In order to test this, we manipulate whether we \emph{learn} the demonstrator model or \emph{assume} it. To avoid confounds introduced by changing the inference algorithm, we use the same algorithm for both. In the learning case, we train the planner on the ground truth demonstrator data; in the assume case, we train it on data generated from a) a Boltzmann-rational demonstrator; and b) an optimal demonstrator -- these are the two models commonly assumed by IRL algorithms. Keeping the algorithm the same enables us to isolate the effect of adapting to an unknown model from the effect of having to use an approximate differentiable planner rather than a perfect one. We will quantify the second effect, i.e. the approximation error introduced by the VIN, in \secref{sec:VI-experiment}.

In the setting where we learn the bias, we further manipulate whether we have access to known rewards for some tasks or not -- i.e. whether we use \algoref{alg:easy} or \algoref{alg:hard}.

Finally, we manipulate the \emph{actual bias} of the demonstrator. Following \citet{IgnorantInconsistentAgents}, we implement the myopic, naive and sophisticated synthetic demonstrators as modifications of the value iteration algorithm. Similarly, we implement the overconfident and underconfident demonstrators by modifying the transition probability distributions used to plan in value iteration. We also include an optimal demonstrator, and stochastic (Boltzmann) versions of all demonstrators.

\prg{Dependent measures.}
We measure the reward obtained by planning optimally with the inferred reward function, as a percentage of the maximum possible reward that could be obtained.

\noindent\textbf{Comparisons among differentiable planners.} Figure~\ref{fig:bias-results} shows the results for learning a demonstrator model vs. assuming an optimal or a Boltzmann demonstrator. The top left subfigure plots what happens on average, across all synthetic demonstrators we tested. The results do provide support to the hypothesis: both learning methods (orange) outperform assuming a model (gray). Looking at the breakdown per demonstrator, we see that assuming optimal does not do well when the demonstrator has any noise (bottom graph). Similarly, assuming Boltzmann does not do well when the demonstrator is not noisy (top graph).

The learning methods tend to perform on par with the best of two choices. In some cases, like the naive and sophisticated hyperbolic discounters, especially the noisy ones, the learning methods outperform both optimal and Boltzmann assumptions. The optimal assumption outperforms the learning methods in some of the non-noisy cases. We hypothesize that this is because as long as the demonstrator eventually reaches the best reward location, assuming optimality allows us to figure out this location. It is then possible to perform near-optimally on our task with knowledge of the best reward location, by navigating to that location and staying there.

Interestingly, Algorithm 1 does not always outperform Algorithm 2, despite it having access to known rewards. We believe this has to do with the fact that Algorithm 2 exploits Assumption 2a (demonstrator close to optimal) and initializes from training on simulated optimal demonstrator data. Algorithm 1 does not rely on this assumption and therefore does not benefit from this initialization, even though the assumption is correct for most of the models we test.

\begin{figure*}
    \centering
    \includegraphics[width=\linewidth]{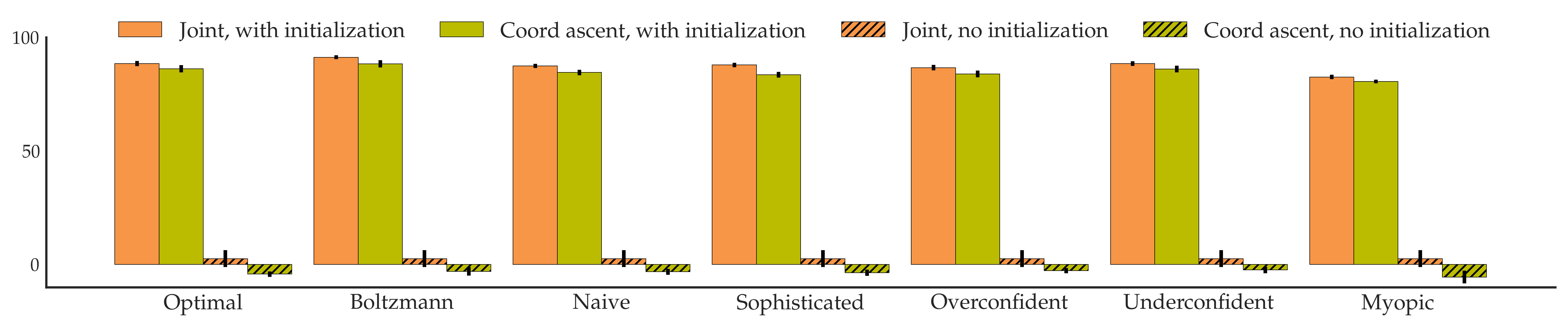}
    \caption{Percent reward obtained for different bias models using variations of Algorithm 2, which does not get access to any known rewards. These algorithms can vary along two dimensions -- whether they are initialized with the assumption that the demonstrator is rational, and whether they train the planner and reward jointly or with coordinate ascent. The original version of Algorithm 2 does initialize, and trains jointly.}
    \label{fig:initialization}
\end{figure*}

\subsection{Tradeoff between being adaptive to bias vs. using exact planning} \label{sec:VI-experiment}
Our paper is about investigating the viability of the "don't assume specific biases" idea. To be adaptive to different kinds of biases the agent might see, it has to learn a model of the demonstrator's planning algorithm via a differentiable planner. Unfortunately, this causes planning to be approximate -- it seems like whatever benefit we get from the adapting to biases, we lose because of the approximation. But these planners will become more practical, they can make this idea practical as well.

To quantify this loss, we replace the VIN with a differentiable exact model of the demonstrator, and infer the reward by backpropagating through the exact model. Since value iteration is not differentiable, we implement soft value iteration, where max operations are replaced with logsumexp operations, and measure percent reward obtained when inferring rewards for an optimal demonstrator.

\noindent\textbf{Results.} With an exact model of the demonstrator, we get $(98.1 \pm 0.1)\%$ of the maximum reward when performing optimal planning on the inferred rewards, while we get $(86.9 \pm 1.6)\%$ with \algoref{alg:easy} and $(86.2 \pm 1.6)\%$ with \algoref{alg:hard}. Again, better planners would improve both algorithms.

\subsection{How important are the various parts of the algorithm?} \label{sec:initialization}

\algoref{alg:hard} was predicated on Assumption 2a, that the demonstrator's planner was "close" to rational, which motivated the initialization step where the planner is trained to mimic an optimal agent. We test how important this is by modifying \algoref{alg:hard} to infer rewards without an initialization (removing lines 1-4). We include versions of the algorithm where we perform coordinate ascent by alternating planner training and reward training instead of training the planner and reward jointly.

\prg{Results.} Figure~\ref{fig:initialization} shows the results for a subset of demonstrators (full results are in the supplementary material). We can see that the initialization is indeed crucial for good performance, as expected. It also turns out that the joint training outperforms coordinate ascent.

\prg{The importance of joint training.} We may also ask what value the joint training adds over the initialization. Without the joint training, Algorithm 2 is simply training the planner to be optimal and then inferring rewards, and so is identical to the Optimal case in Figure~\ref{fig:bias-results}. We can see that the joint training does add value over the initialization.

\section{Discussion}

\prg{Summary.} It seems daunting to try to characterize human biases, and yet assuming the wrong bias can lead to agents that do not correctly understand what people want. A natural alternative is to let the data characterize the bias: to learn how people generate their actions given what they want, with all the biases they might have. Our goal in this work was to investigate this approach and gain insight into whether it just works right off the bat, and, if not, what additional structure it benefits from and what improvements in state of the art algorithms it requires. Overall, we have found that it might be possible to maintain flexibility in learning systematic biases while regularizing the learned planner to be close to optimal; but also that for this to be practical within a deep learning architecture will need more progress in differentiable planning. 


\prg{Limitations and future work.} A core limitation of our analysis is that it was conducted on simple problems in simulation, rather than on real problems with real people. This makes sense as a starting point, because it allows us to have ground truth to evaluate against, and allows us to generate large datasets that would be hard to collect with real humans. However, as differentiable planners are able to handle higher complexity, the analysis ought to eventually move to more realistic tasks with human data.


Further, we did not explore all possible assumptions that could simplify the planner learning task. The assumption that the demonstrator has the same bias across many tasks is key to the algorithms, but is very strong. This analysis could be extended by using meta-learning to learn a prior over planners, and by inferring the demonstrator's beliefs as in \citet{BayesianTOM} (e.g. via TOMNets \citep{MachineTOM}). We are excited to look into this, and into what additional inductive bias we could leverage, in our future work.

\subsubsection*{Acknowledgments}

We thank the researchers at the Center for Human Compatible AI for valuable feedback. This work was supported by the Open Philanthropy Project, AFOSR, and National Science Foundation Graduate Research Fellowship Grant No. DGE 1752814.

\bibliography{references}
\bibliographystyle{icml2019}

\newpage

\ 

\newpage

\appendix

\section{Additional experimental data}

In \secref{sec:evaluation}, we presented data on the results of running various algorithms against a set of demonstrators, reporting the reward obtained according to the true reward function when using the inferred reward with an optimal planner, as a percentage of the maximum possible true reward. \tblref{table:reward} shows the percentage reward obtained for all combinations of algorithms and demonstrators. We also measure the accuracy of the planner and reward at predicting the demonstrator's actions in new gridworlds where the rewards are the same but the wall locations have changed. These results are presented in \tblref{table:accuracy}. Note that there are often multiple optimal actions at a given state, which makes it challenging to get high accuracy.

\begin{table*}[hb]
  \caption{Percent reward obtained when the algorithm (column) is used to infer the bias of the demonstrator (row). The optimal and Boltzmann algorithms assume a fixed model of the demonstrator and train the VIN to mimic the model before performing reward inference (and were used in \figref{fig:bias-results}). We also include the four flavors of \algoref{alg:hard} that were plotted in \figref{fig:initialization}. The VI algorithm uses a differentiable implementation of soft value iteration as the planner instead of a VIN (used in \secref{sec:VI-experiment}). The demonstrators are the optimal agent, the biased agents of \figref{fig:biases}, and versions of each of these agents with Boltzmann noise.}
  \label{table:reward}
  \centering
  \resizebox{\linewidth}{!}{
  \begin{tabular}{c | *{8}{c}}
    \toprule
    \cmidrule{1-2}
    Agent & Optimal & Boltzmann & Algorithm 1 & Coord w/ init & Joint w/ init & Coord w/o init & Joint w/o init & VI \\
    \midrule
    Average & $ 67.0 \pm 2.7 $ & $ 79.7 \pm 0.8 $ & $ 89.5 \pm 0.7 $ & $ 85.0 \pm 0.5 $ & $ 86.4 \pm 0.6 $ & $ -3.9 \pm 0.7 $ & $ 2.6 \pm 1.0 $ & $ 71.9 \pm 3.0$ \\
    Optimal & $ 87.3 \pm 1.0 $ & $ 73.9 \pm 2.3 $ & $ 86.9 \pm 1.6 $ & $ 86.2 \pm 1.6 $ & $ 88.5 \pm 1.1 $ & $ -4.2 \pm 1.2 $ & $ 2.6 \pm 3.7 $ & $ 98.1 \pm 0.1$ \\
    Naive & $ 86.4 \pm 0.9 $ & $ 74.4 \pm 1.6 $ & $ 91.1 \pm 0.8 $ & $ 84.6 \pm 1.2 $ & $ 87.5 \pm 0.9 $ & $ -3.2 \pm 1.3 $ & $ 2.6 \pm 3.7 $ & $ 96.1 \pm 0.1 $ \\
    Sophisticated & $ 87.5 \pm 1.1 $ & $ 77.1 \pm 1.6 $ & $ 91.8 \pm 1.3 $ & $ 83.6 \pm 1.3 $ & $ 87.9 \pm 1.0 $ & $ -3.6 \pm 1.4 $ & $ 2.6 \pm 3.7 $ & $ 96.7 \pm 0.1 $ \\
    Myopic & $ 82.8 \pm 0.8 $ & $ 77.0 \pm 1.2 $ & $ 81.0 \pm 2.8 $ & $ 80.6 \pm 0.8 $ & $ 82.6 \pm 1.0 $ & $ -5.5 \pm 2.8 $ & $ 2.6 \pm 3.7 $ & $ 87.5 \pm 0.2 $ \\
    Overconfident & $ 87.5 \pm 1.2 $ & $ 70.7 \pm 1.7 $ & $ 82.1 \pm 1.4 $ & $ 83.9 \pm 1.5 $ & $ 86.7 \pm 1.2 $ & $ -2.7 \pm 1.1 $ & $ 2.6 \pm 3.7 $ & $ 97.5 \pm 0.1 $ \\
    Underconfident & $ 88.0 \pm 0.8 $ & $ 74.7 \pm 1.6 $ & $ 86.7 \pm 1.2 $ & $ 86.1 \pm 1.5 $ & $ 88.5 \pm 1.0 $ & $ -2.4 \pm 1.4 $ & $ 2.6 \pm 3.7 $ & $ 98.9 \pm 0.2 $ \\
    Boltzmann & $ 8.5 \pm 1.0 $ & $ 90.7 \pm 1.3 $ & $ 91.4 \pm 0.8 $ & $ 88.4 \pm 1.6 $ & $ 91.3 \pm 0.9 $ & $ -3.0 \pm 1.9 $ & $ 2.6 \pm 3.7 $ & $ 8.7 \pm 0.1 $ \\
    B-Naive & $ 52.8 \pm 2.3 $ & $ 77.3 \pm 2.9 $ & $ 98.5 \pm 0.1 $ & $ 82.5 \pm 2.4 $ & $ 75.8 \pm 2.9 $ & $ -8.3 \pm 4.5 $ & $ 2.6 \pm 3.7 $ & $ 47.7 \pm 0.2 $ \\
    B-Sophisticated & $ 51.5 \pm 2.1 $ & $ 74.5 \pm 2.8 $ & $ 98.8 \pm 0.2 $ & $ 80.1 \pm 1.5 $ & $ 77.0 \pm 2.3 $ & $ -8.7 \pm 3.9 $ & $ 2.6 \pm 3.7 $ & $ 48.0 \pm 0.2 $ \\
    B-Myopic & $ 77.7 \pm 1.1 $ & $ 90.8 \pm 0.6 $ & $ 95.6 \pm 1.0 $ & $ 91.5 \pm 0.6 $ & $ 91.9 \pm 0.5 $ & $ -2.4 \pm 2.1 $ & $ 2.6 \pm 3.7 $ & $ 83.4 \pm 0.1 $ \\
    B-Overconfident & $ 7.0 \pm 0.9 $ & $ 84.1 \pm 2.3 $ & $ 79.2 \pm 2.3 $ & $ 81.4 \pm 2.8 $ & $ 86.3 \pm 1.3 $ & $ -0.8 \pm 1.6 $ & $ 2.6 \pm 3.7 $ & $ 8.7 \pm 0.1 $ \\
    B-Underconfident & $ 86.7 \pm 0.9 $ & $ 91.3 \pm 0.7 $ & $ 91.2 \pm 0.7 $ & $ 90.7 \pm 1.0 $ & $ 92.4 \pm 0.8 $ & $ -1.8 \pm 1.2 $ & $ 2.6 \pm 3.7 $ & $ 92.1 \pm 0.1 $ \\
    \midrule
  \end{tabular}}
\end{table*}

\begin{table*}[hb]
  \caption{Accuracy when predicting the demonstrator's actions (row) on new gridworlds using the planner and reward inferred by the algorithm (column). Algorithms and demonstrators are the same as in \tblref{table:reward}.}
  \label{table:accuracy}
  \centering
  \resizebox{\linewidth}{!}{\begin{tabular}{c | *{8}{c}}
    \toprule
    \cmidrule{1-2}
    Agent & Optimal & Boltzmann & Algorithm 1 & Coord w/ init & Joint w/ init & Coord w/o init & Joint w/o init & VI \\
    \midrule
    Optimal & $ 61.3 \pm 0.4 $ & $ 59.8 \pm 0.4 $ & $ 62.0 \pm 0.3 $ & $ 62.8 \pm 0.2 $ & $ 63.6 \pm 0.3 $ & $ 63.0 \pm 0.2 $ & $ 72.4 \pm 0.1 $ & $ 25.7 \pm 0.1 $ \\
    Naive & $ 60.1 \pm 0.3 $ & $ 59.4 \pm 0.3 $ & $ 58.6 \pm 0.3 $ & $ 61.3 \pm 0.3 $ & $ 61.8 \pm 0.3 $ & $ 61.0 \pm 0.3 $ & $ 71.1 \pm 0.1 $ & $ 24.9 \pm 0.1 $ \\
    Sophisticated & $ 60.5 \pm 0.4 $ & $ 59.2 \pm 0.4 $ & $ 59.3 \pm 0.3 $ & $ 61.0 \pm 0.3 $ & $ 62.0 \pm 0.4 $ & $ 61.2 \pm 0.3 $ & $ 71.2 \pm 0.1 $ & $ 24.9 \pm 0.1 $ \\
    Myopic & $ 54.1 \pm 0.4 $ & $ 53.5 \pm 0.5 $ & $ 54.9 \pm 0.5 $ & $ 55.6 \pm 0.2 $ & $ 56.1 \pm 0.3 $ & $ 56.0 \pm 0.1 $ & $ 62.8 \pm 0.1 $ & $ 20.4 \pm 0.1 $ \\
    Overconfident & $ 61.6 \pm 0.4 $ & $ 60.1 \pm 0.4 $ & $ 61.8 \pm 0.4 $ & $ 63.3 \pm 0.3 $ & $ 63.7 \pm 0.3 $ & $ 63.1 \pm 0.2 $ & $ 72.8 \pm 0.1 $ & $ 25.9 \pm 0.1 $ \\
    Underconfident & $ 60.9 \pm 0.4 $ & $ 59.5 \pm 0.4 $ & $ 61.4 \pm 0.3 $ & $ 62.4 \pm 0.3 $ & $ 62.9 \pm 0.3 $ & $ 62.5 \pm 0.3 $ & $ 72.0 \pm 0.1 $ & $ 25.5 \pm 0.1 $ \\
    Boltzmann & $ 56.7 \pm 1.1 $ & $ 60.5 \pm 0.4 $ & $ 60.9 \pm 0.3 $ & $ 60.3 \pm 0.2 $ & $ 60.8 \pm 0.3 $ & $ 62.3 \pm 0.3 $ & $ 67.1 \pm 0.5 $ & $ 24.2 \pm 0.1 $ \\
    B-Naive & $ 56.6 \pm 0.8 $ & $ 59.8 \pm 0.8 $ & $ 60.4 \pm 0.1 $ & $ 60.3 \pm 0.2 $ & $ 60.5 \pm 0.7 $ & $ 59.9 \pm 0.3 $ & $ 68.5 \pm 0.3 $ & $ 23.7 \pm 0.1 $ \\
    B-Sophisticated & $ 57.6 \pm 0.7 $ & $ 60.2 \pm 0.7 $ & $ 60.5 \pm 0.2 $ & $ 60.5 \pm 0.2 $ & $ 61.2 \pm 0.3 $ & $ 60.1 \pm 0.3 $ & $ 68.5 \pm 0.3 $ & $ 23.7 \pm 0.1 $ \\
    B-Myopic & $ 56.3 \pm 0.2 $ & $ 56.9 \pm 0.4 $ & $ 55.9 \pm 0.2 $ & $ 56.5 \pm 0.2 $ & $ 57.0 \pm 0.2 $ & $ 56.3 \pm 0.1 $ & $ 62.4 \pm 0.1 $ & $ 20.3 \pm 0.0 $ \\
    B-Overconfident & $ 56.9 \pm 1.1 $ & $ 60.7 \pm 0.4 $ & $ 61.3 \pm 0.3 $ & $ 60.9 \pm 0.2 $ & $ 61.6 \pm 0.3 $ & $ 62.7 \pm 0.2 $ & $ 68.0 \pm 0.5 $ & $ 24.2 \pm 0.1 $ \\
    B-Underconfident & $ 62.4 \pm 0.3 $ & $ 63.1 \pm 0.4 $ & $ 63.4 \pm 0.2 $ & $ 63.0 \pm 0.1 $ & $ 63.6 \pm 0.1 $ & $ 63.5 \pm 0.2 $ & $ 72.2 \pm 0.1 $ & $ 25.4 \pm 0.1 $ \\
    \bottomrule
  \end{tabular}}
\end{table*}

\end{document}